\documentclass{article} 
\usepackage{iclr2022_conference,times}

\usepackage{amsmath,amsfonts,bm}

\def\eqref#1{equation~\ref{#1}}

\def\1{\bm{1}}

\DeclareMathAlphabet{\mathsfit}{\encodingdefault}{\sfdefault}{m}{sl}
\SetMathAlphabet{\mathsfit}{bold}{\encodingdefault}{\sfdefault}{bx}{n}

\DeclareMathOperator*{\argmin}{arg\,min}

\usepackage{hyperref}
\usepackage{url}

\usepackage{graphicx}
\usepackage{subfigure}
\newcommand{\matr}[1]{\mathbf{#1}}

\newcommand{\X}{\mathbf{X}}
\newcommand{\T}{\mathbf{T}}
\newcommand{\Xflip}{\mathbf{X_{f}}}
\newcommand{\Y}{\mathbf{Y}}
\newcommand{\Yflip}{\mathbf{Y_{f}}}

\title{Learning Canonical Embedding for Non-rigid Shape Matching}

\author{Abhishek Sharma \& Maks Ovsjanikov  \\
LIX, Ecole Polytechnique\\
IP Paris, France\\
\texttt{\{sharma,maks\}@@lix.polytechnique.fr } \\
}

\iclrfinalcopy 
\begin{document}

\maketitle

\begin{abstract}
This paper provides a novel framework that learns canonical embeddings for non-rigid shape matching. In contrast to prior work in this direction, our framework is trained end-to-end and thus avoids instabilities and constraints associated with the commonly-used Laplace-Beltrami basis or sequential optimization schemes. On multiple datasets, we demonstrate that learning self symmetry maps with a deep functional map projects 3D shapes into a low dimensional canonical embedding that facilitates non-rigid shape correspondence via a simple nearest neighbor search. Our framework outperforms multiple recent learning based methods on FAUST and SHREC benchmarks while being computationally cheaper, data efficient, and robust.
\end{abstract}

\section{Introduction}
\label{sec:intro}
Shape correspondence is a fundamental problem in computer vision, computer graphics and related fields \citep{Thomas21}, since it facilitates  many applications such as texture or deformation transfer and statistical shape analysis ~\citep{bogo2014} to name a few. Although shape correspondence has been studied from many viewpoints, we focus here on a functional map-based approaches \citep{ovsjanikov2012functional} as this framework is quite general, scalable and thus, has been extended to various other applications such as pose estimation \citep{neverova20}, matrix completion \citep{sharmamatrix} and graph matching \citep{FRGM20}.

While recent learning based deep functional map approaches have made impressive gains in non rigid isometric full shape matching \citep{litany17,roufosse2019unsupervised,halimi2018self,sharma20}, partial shape matching \citep{rodola2017partial, litany17} has received little attention despite it being of great interest in robotics \citep{ijrr12} and Virtual reality applications \citep{sharma16}. The progress is mainly hindered by the difficulty of learning a suitable embedding or basis functions for partial 3D data. The majority of works in this domain use the Laplace-Beltrami basis ~\citep{ovsjanikov2017computing}, which are biased towards near-isometries and can be unstable under significant partiality \citep{kirgo2020wavelet}. As a consequence, there is no unified framework that excels at learning both partial as well as full non-rigid shape matching despite some efforts in this direction \citep{sharma20,ric_linear20}. While the former approach \cite{sharma20} is class-specific and requires retraining a network for each class, the latter \citep{ric_linear20} employs a two stage optimization strategy without adequate regularization which, as we show later, is suboptimal.

Instead of using predefined basis functions in the functional map framework, learning an embedding, to be used as a basis in the functional map, is a promising direction towards obtaining a unified shape matching framework for both partial as well as full shape matching. However, there exist few works that exploit  prior information on such embeddings. \citet{ric_linear20} make the first attempt by assuming such embeddings to be linearly invariant between a pair of shapes. However, as we show later, learning such an embedding without exploiting natural priors on 3D shapes, such as their symmetry structure, leads to overfitting as no regularization or constraint is enforced on the linear transformation between a pair of shape embeddings \citep{ric_linear20}.

In this paper, we present a novel non-rigid shape matching method based on a nearest neighbour approach in canonical embedding. We work with point cloud representation of 3D shapes and assume to be given a self symmetry map for each shape during training. We make two assumptions on the canonical embedding: our first assumption is to learn an embedding of each shape that would make the given self-symmetry map \textit{linear} in some higher-dimensional space. This is advantageous for two reasons: first, it significantly simplifies the embedding learning pipeline and makes it learnable end-to-end. This is because the linearly invariant assumption is made on a self-symmetry map and not on the pairwise map between shapes as done in \citep{ric_linear20}. This alleviates the need to retrieve this linear transformation for shape matching at test time and thereby, reduces the two stage sequential optimization scheme of \cite{ric_linear20} to joint optimization. Secondly, modelling a self symmetry map enables us to explicitly enforce the pointwise map between two shapes to take into account the intrinsic self-symmetry during training. Thus, our second contribution is a novel regularizer that, combined with the learned self symmetry map during training, significantly improves generalization and robustness to sampling resolution as well as the size of embedding. Our method obtains superior results on multiple shape matching benchmarks such as FAUST and SHREC when compared to recent learning-based methods while being computationally cheaper, more robust and data efficient.


\section{Related Work}
\label{sec:related}

\paragraph{Functional Maps} Computing point-to-point maps between two 3D discrete surfaces is a very well-studied problem. We refer to a recent survey \citep{sahilliouglu2019recent}  for an in-depth discussion. Our method is closely related to the functional map pipeline, introduced in \citep{ovsjanikov2012functional} and then significantly extended in follow-up works (see, e.g.,\citet{ovsjanikov2017computing}). The key idea of this framework is to encode correspondences as small matrices, by using a reduced functional basis, thus greatly simplifying many resulting  optimization problems. The functional map pipeline has been further improved in accuracy, efficiency and robustness by many recent works including ~\citep{kovnatsky2013coupled,huang2014functional,burghard2017embedding,rodola2017partial,commutativity,ren2018continuous,smoothshells,CyclicFM20}. There also exist other works \citep{wei2016dense,MasBosBroVan16,monti2017} that treat shape correspondence as a dense labeling problem but they typically require a lot of data as the label space is very large.

\paragraph{Learning from raw 3D shape}
Although early approaches in functional maps literature used hand-crafted features \citep{ovsjanikov2017computing}, more recent methods directly aim to \emph{learn} either the optimal transformations of hand crafted descriptors \citep{litany17,roufosse2019unsupervised} or even  features directly from 3D geometry itself \citep{donati20,sharma20}. Initial efforts in this direction used classical optimisation techniques \citep{corman2014supervised}. In contrast, Deep Functional Maps \citep{litany2017deep} proposed a deep learning architecture called FMNet to optimize a non-linear transformation of SHOT descriptors \citep{shot}, that was further extended to unsupervised setting \citep{roufosse2019unsupervised,halimi2018self}. To alleviate the sensitivity to SHOT descriptor, recent works including \citep{groueix20183d,donati20,sharma20} learn shape matching directly from the \emph{raw 3D data} without relying on pre-defined descriptors, thus leading to improvements in both robustness and accuracy. However, all these works are aimed at \textit{full} (complete) shape correspondence and do not handle partial shape matching effectively.

\paragraph{Self Supervised Learning} Self supervised learning has been exploited for learning representations and embedding in various domains where a proxy task is used to learn the representation. e.g. \cite{sharma16} uses an autoencoder to complete the partial shapes and uses the resulting representation of shape completion for shape classification task. \cite{rot18_komo} learns to predict image rotations and uses the resulting representation for image classification. Our formulation is in the same spirit as we learn to inject the symmetry information in a 3D shape and use the resulting representation for 3D shape matching. However, we choose symmetry learning as a proxy task for embedding learning for a principled reason which we describe in detail in the methodology section.

\paragraph{Learning Basis from Data} Most of the functional map frameworks can not handle partiality in data as they rely on Laplacian eigenfunctions that are shown to be unstable under partial data. \citep{rodola2017partial,litany17,pfm_new20} deal with partiality but they are based on hand-crafted features and require an expensive optimization scheme and are instance specific. While \cite{sharma20} proposes to learn a suitable alignment of pre-computed Laplacian Eigen basis functions, the approach still relies on the Laplacian basis and can therefore be unstable. \cite{ric_linear20} proposed a two stage architecture to learn a linear transformation invariant shape embedding to bypass the difficulties associated with LBO. However, as we  demonstrate later in experiments, the two stage architecture is suboptimal due to the lack of adequate regularization.
\paragraph{Symmetry for Non Rigid Shape Matching} Matching shapes with intrinsic symmetries involves dealing with symmetric ambiguity problem which has been very well studied and explored in axiomatic methods~\citep{Raviv10,sym_Lipman10,mitrastar12,maks_quotient13,jing_sig20}. More recently, \citep{sym-img,clara20} proposes an end to end method to learn extrinsic 3D symmetries from a RGB-D image. However, none of the existing learning based non-rigid shape matching method models or learn symmetry explicitly as a regularizer for shape matching. 



 Rest of the paper is structured as follows: In the next section, we briefly cover the necessary background on the functional map. Afterwards, we propose a novel learning strategy to learn canonical embedding and introduce our novel regularization based on intrinsic symmetry prior. Lastly, we validate our framework on three benchmark datasets by comparing it to various state-of-the-art methods and providing ablation studies.
 
 \section{Background}
 Before describing our method, we provide a brief overview of the basic pipeline to compute a functional map \citep{ovsjanikov2012functional}. 

\paragraph{Functional Map Computation } The typical functional map pipeline  \citep{ovsjanikov2012functional} assumes that we are given a source and a target shape, $\X,\Y$, containing, respectively, $n_x$ and $n_y$ vertices, a small set of $k$ basis functions, e.g. of the respective Laplace-Beltrami operators (LBO). We are also given a set of descriptors on each shape, to be preserved by the unknown map, whose coefficients in the basis functions are stored as columns of matrices $\Phi_{\X}, \Phi_{\Y}$. The optimal \emph{functional map} $\matr{C}_{\text{opt}}$ is computed by solving the following optimization problem:
\begin{align}
\label{eq:opt_problem}
\matr{C}_{\text{opt}} = \argmin_{\matr{C}} E_{\text{desc}}\big(\matr{C}\big) + \alpha E_{\text{reg}}\big(\matr{C}\big),
\end{align}
where  $E_{\text{desc}}\big(\matr{C}\big)
= \big\Vert \matr{C} \Phi_{\X} - \Phi_{\Y}\big\Vert^2$ aims at the descriptor preservation whereas the second term acts as a regularizer on the map by enforcing its overall structural properties, such as bijectivity of the map. The optimization problem in \eqref{eq:opt_problem} can be solved with any convex solver. Once the optimal functional map $\matr{C}_{\text{opt}}$ is computed, one can use nearest neighbor search in the spectral embedding to convert it to a point to point correspondence.

 Note that when  the basis functions are neural network-based, instead of optimizing over $\mathbf{C}$, we are optimizing the functional in \eqref{eq:opt_problem}  over $\mathbf{C}$, $\Phi_{\X}$ and $\Phi_{\Y}$. In this case, joint optimization over $\mathbf{C}$, $\Phi_{\X}$ and $\Phi_{\Y}$ is challenging as $\mathbf{C}$ is computed via an iterative solver itself. 


\section{Learning Canonical Embedding}
\label{sec:method}
In the previous section, we outlined a basic mechanism to compute a functional map given a set of basis functions. Due to the instability of Laplace-Beltrami operator on partial 3D shapes, our main goal is to avoid using its eigenfunctions and instead aim to \emph{learn} a embedding that can replace the spectral embedding given by the LBO. This section details how to learn such an embedding whilst working in the symmetric space.

\textbf{Input Shape Representation} In contrast to several recent works \citep{halimi2018self,sharma20} that assume to be given a mesh representation of 3D shapes in non-rigid shape matching, we do not impose any such constraint and directly work the with point cloud representation.

\textbf{Type of Supervision} In addition to the pointwise map between shapes, we assume to be given a collection of shapes with a self-symmetry ground truth map for each shape and our goal is to find an embedding that respects the given symmetry of each shape and that ultimately can reduce shape correspondence between a pair of shapes to a nearest neighbor search between their embeddings. When compared to our main baseline \citep{ric_linear20}, this is an additional supervision and thus, we also evaluate \cite{ric_linear20} with the same supervision for fair comparison. 
 
 Our work is most closely related to a recent work \citep{ric_linear20} that proposes to replace the Laplace-Beltrami basis by learning embeddings that are related by a linear transformation across pairs of shapes. Intuitively, this formulation aims to embed a shape from the 3D space, in which complex non-rigid deformations could occur, to another higher-dimensional space, in which transformations across shapes are linear. However, using a supervised loss to learn this embedding  without enforcing any structural properties on the underlying linear transform provides little guarantee that the learned transform will generalize from the train to test setting.

 \paragraph{Linearly Invariant self-symmetry embedding}  We denote a map between a pair of shapes $\X$ and $\Y$ by $T_{\X\Y}:\X \rightarrow \Y$ such that $T_{\X\Y}(x_i) = y_j$, $\forall i \in \{1, \ldots , n_{\X}\}$ and some $j  \in \{1, \ldots , n_{\Y}\}$.  This map can be represented by a matrix $\Pi_{\X\Y} \in \mathbb{R}^{n_{\X}\times n_{\Y}}$ such that $\Pi_{\X\Y}(i,j) = 1$ if  $T_{\X\Y}(x_i) = y_j$ and $0$ otherwise. We use the same notation $T$ for self symmetry map $\T_{\X\Xflip}$ as well.  We use $P_{\X}$ to denote the 3D coordinates of $\X$. 
 
\begin{figure}[h]
   \begin{minipage}{0.48\textwidth}
        \centering
        
        \includegraphics[width=.94\linewidth]{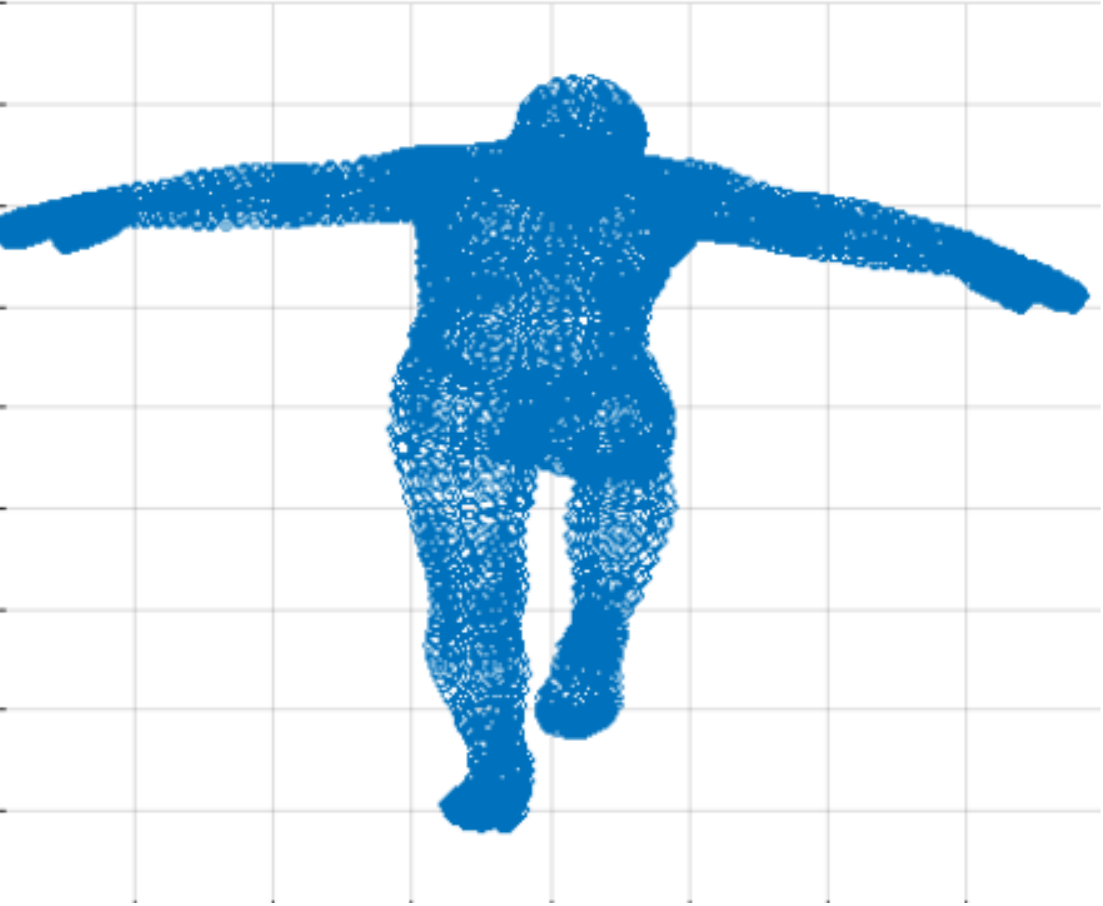}
    \end{minipage}%
    \begin{minipage}{0.48\textwidth}
       \centering
        \includegraphics[width=.94\linewidth]{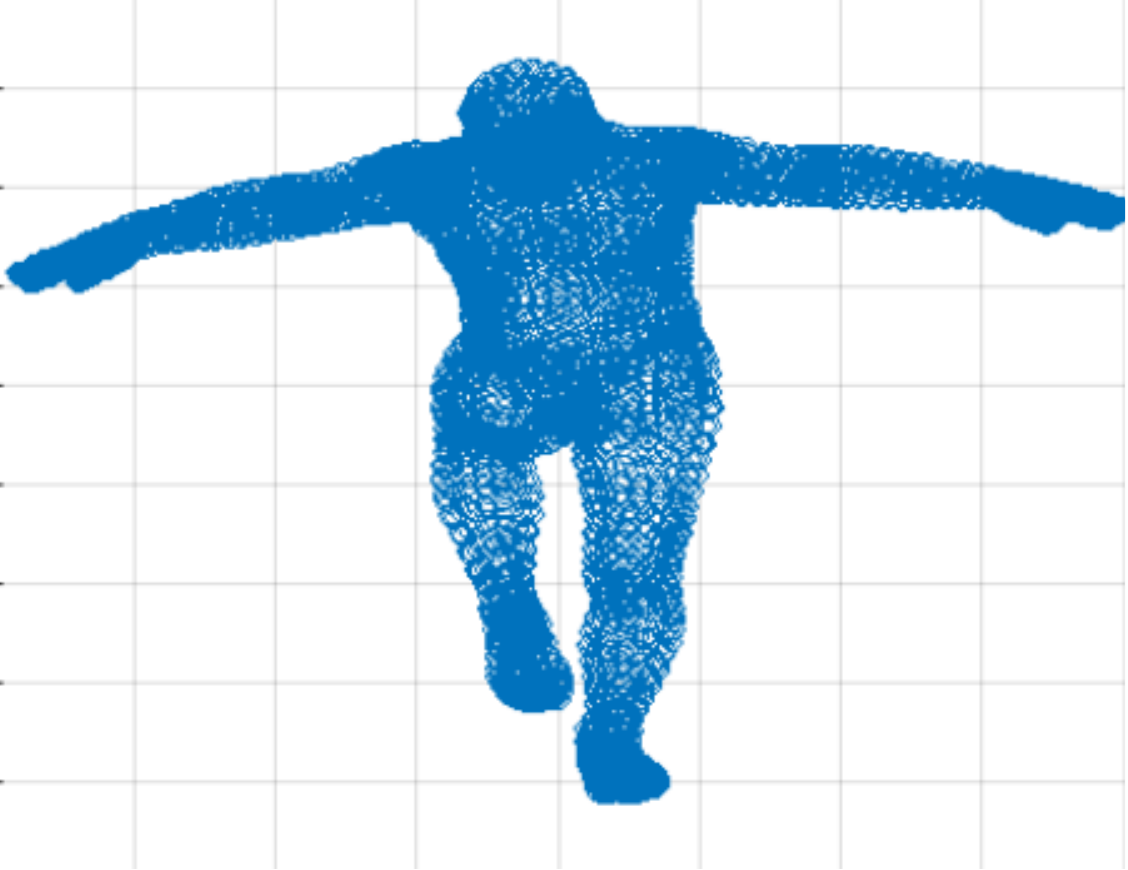} 
    \end{minipage}
    \caption{Original Shape and its Flipped version from Surreal dataset}
    \label{fig:flip}
\end{figure}

 Our network takes a shape $\X$ as input as well as its point to point symmetry map denoted as $\T_{\X\Xflip}$. We then perform a reflection (flip) of each shape along one axis resulting in a shape denoted as $\Xflip$. The original and flipped shapes are then forwarded to a Siamese architecture, based on a PointNet~\cite{qi2017pointnet} feature extractor,  that embeds these two shapes into some fixed $k$ dimensional space. We illustrate in Figure \ref{fig:flip} one such flip. The intuition behind this operation is to help the network learn representation that can disambiguate left from right in shape matching. Our first key idea is to learn an embedding of each shape that would make the given self-symmetry map linear in some higher-dimensional space. 

 Let $\Phi_{\X}$ and $\Phi_{\Xflip}$ denote the matrices, whose rows can be interpreted as embeddings of the points of $\X$ and $\Xflip$. In the functional map framework, there exists a functional map  $C_{\X\Xflip}$ that aligns the corresponding embeddings. Given a self symmetry ground truth pointwise map $\T_{\X\Xflip}$, we can estimate $C_{\X\Xflip}$ by solving the following optimization problem: 
 \begin{equation}
     C_{\X\Xflip} = \argmin_{C} \|\Phi_{\X}  C^T  - \T_{\X\Xflip} \Phi_{\Xflip} \|_2  
 \end{equation}

 The optimal symmetry map $C_{\X\Xflip}$ is given by: $C_{\X\Xflip} = (\Phi_{\X}^{+} \T_{\X\Xflip} \Phi_{\Xflip})^T$, that is differentiable using the closed-form expression of derivatives of matrix inverses, as also mentioned in Section $3$. Similarly, we can compute 
$C_{\Y\Yflip}$ for shape $\Y$.

\subsection{Loss functions}
Given a set of pairs of shapes $\X,\Y$ for which ground truth correspondences $\T^{gt}_{\X\Y}$ are known along with a pointwise symmetry map, our network  computes an embedding  $\Phi_{\X}, \Phi_{\Y}$ for each shape as well as a self symmetry functional map $C_{\X\Xflip}$ and $C_{\Y\Yflip}$ respectively as described above. We then optimize the sum of three loss functions, one each defined for linearly invariant self symmetry embedding, nearest neighbour based loss for pairwise (shape pair) embedding and a commutativity loss for explicitly enforcing intrinsic symmetry during training. 

\paragraph{Linearly Invariant Loss} The first two loss functions are based on a soft-correspondence matrix, also used in \citep{litany2017deep} and \citep{ric_linear20}. To define it for self symmetry map, we \emph{transform} each shape embedding $\widehat{\Phi}_{\X} = \Phi_{\X} C_{\X\Xflip}^T$ by applying the optimal symmetry map. We then compare the rows of $\widehat{\Phi}_{\X}$ to those of $\Phi_{\Xflip}$ to obtain the \textit{soft} correspondence matrix $S_{\X\Xflip}$ that approximates the self-symmetry map in a differentiable way as follows:
\begin{equation}
    \begin{aligned}
        (S_{\X\Xflip})_{ij} = \frac{e^{- \|\widehat{\Phi}_{\X}^{i} - \Phi_{\Xflip}^{j}\|_2 }}{\sum_{k=1}^n e^{- \|\widehat{\Phi}_{\X}^{i} - \Phi_{\Xflip}^{k}\|_2 }} 
    \end{aligned}
    \label{eq:softmax}
\end{equation}

We then define our loss that uses this soft-map to transfer the Euclidean coordinates and compares the result to transferring the coordinates using the ground truth map.
\begin{equation}
\begin{split}
 L(\Phi_{\X}, \Phi_{\Xflip},\Phi_{\Y}, \Phi_{\Yflip})_{lin.} =  \sum \|  S_{\X\Xflip} P_{\Xflip} - \T^{gt}_{\X\Xflip} P_{\Xflip} \|_2^2 + \sum \|  S_{\Y\Yflip} P_{\Yflip} - \T^{gt}_{\Y\Yflip} P_{\Yflip} \|_2^2 \\
    \end{split}
    \label{eq:loss_spatial}
\end{equation}
Note that this does not assume that the Euclidean coordinates to correspond. Instead, this loss measures how well the predicted map transfers a particular set of functions, compared to the ground truth map. This loss was introduced in \cite{ric_linear20} but we enforce it on the self-symmetry map.

\paragraph{Euclidean Loss}
The loss described in the previous paragraph only considers the embedding of each shape independently and aims to promote the structural property of this embedding: i.e., that the symmetry map should be linear in the embedding space.

Our next loss links the embeddings of the two shapes and is designed to preserve the given ground truth mapping. Specifically, we first compute the soft correspondence matrix $S_{\X\Y}$ between a pair of shapes, by comparing the rows of $\Phi_{\X}$ to those of $\Phi_{\Y}$ in a differentiable way as done in \eqref{eq:softmax}. We then evaluate the computed soft map, again, by evaluating how well it transfers the coordinate functions, compared to the given ground truth mapping.
\begin{equation}
L(\Phi_{\X},\Phi_{\Y})_{euc.} = \sum \|  S_{\X\Y} P_{\Y} - \T^{gt}_{\X\Y} P_{\Y} \|_2^2.
\label{eq:loss_euc}
\end{equation}
Note that unlike the linearly invariant loss that we impose on the symmetry maps, this loss is based on comparing $\Phi_{\X}$ and $\Phi_{\Y}$ directly, without computing any linear transformations. This significantly simplifies the learning process and in particular, reduces the computation of the correspondence at test time to a simple nearest-neighbor search. Despite this, as we show below, due to our strong regularization, our approach achieves superior results compared to the method of \cite{ric_linear20}, based on computing an optimal linear transformation at test time.

\paragraph{Symmetry Commutativity Loss} Our final loss aims to link the symmetry map computed for each shape and the correspondence across the two shapes. We achieve this by using the algebraic properties of the functional representation, and especially using the fact that map composition can simply be expressed as matrix multiplication.

Specifically, given a self-symmetry pointwise map on shapes $\X$ and shape $\Y$, we aim to promote the \textit{consistency} between the computed correspondence and the symmetries on each shape. We do this by imposing the following commutativity loss during training:
\begin{equation}
     L(\Phi_{\X},\Phi_{\Y})_{comm.} =  \|S_{\Yflip\Y} S_{\X\Y} - S_{\X\Y} S_{\Xflip\X} \|_2
     \label{eq:comm}
 \end{equation}

Intuitively, this loss considers the difference between mapping from $\X$ to $\Y$ and applying the symmetry map on $\Y$, as opposed to applying the symmetry on $\X$ and then mapping from $\X$ to $\Y$. Note that this is similar to the commonly used \textit{Laplacian} commutativity in the functional maps literature. However, rather than promoting isometries, our loss enforces that the computed map respects the self-symmetry structure of each shape, which holds regardless of the deformation class, and is not limited to isometries.
 
 \paragraph{Overall training Loss} We combine the two embedding losses defined in \eqref{eq:loss_spatial} and \eqref{eq:loss_euc} with that of commutativity loss defined in \eqref{eq:comm} and define the training loss as follows:
 \begin{equation}
     L_{tot.} =  L_{euc.} + \lambda *L_{lin.} + \gamma *L_{comm.} 
     \label{eq:tot}
 \end{equation}

The scalars $\lambda$ and $\gamma$ allow us to weigh the symmetry information differently in partial and full shape matching. Naturally, we set them higher for full shape matching where enforcing symmetry structure makes more sense than partial setting where symmetry is partial at times. We set $\lambda$ and $\gamma$ to $5$ for full shape matching and $1$ and $.1$ for partial shape matching based on a small validation set.
\paragraph{Test Phase} At test time, once the network is trained, we simply compute the embedding $\Phi_{\X}$ and $\Phi_{\Y}$ and do a nearest neighbour search between them to find correspondence between the two shapes.

\paragraph{Implementation Details}
We implemented our method in Pytorch \cite{pytorch}. All experiments are run on a Nvidia RTX $2080$ graphics processing card and require $16$ GB of GPU memory. For our experiments, similar to prior work \citep{sharma20,ric_linear20}, we train over randomly selected $5000$ shapes from the SURREAL dataset \citep{varol17_surreal}, where each point cloud is resampled randomly at $3$K vertices. We learn a $k=50$ dimensional embedding (basis) for each point cloud. 

During training, we require self-symmetry ground truth as well as pairwise ground truth map. Following \citep{sharma20,ric_linear20}, our feature extractor is also based on the semantic segmentation architecture of PointNet. We use a batch size of $20$ and learning rate of $1e-4$ and optimize our objective with Adam optimizer in Pytorch \cite{pytorch}. Training iterations as well as other hyperparameters are validated on a small validation set of $500$ shapes from the SURREAL dataset. Unlike \cite{ric_linear20} that uses fixed $1000$ points during train and test time, we use full resolution at test time. In the SURREAL dataset, all point cloud contain $6890$ points. During training, we randomly sample $3000$ points from the point cloud and also augment the training set with random rotations applied to input data. We obtain an embedding of $50$ dimensions during training by validating from the set $40,50,60$. Our results are not sensitive to small changes in these two parameters. We  probe the effect of changing size of embedding as well as the amount of training data on the resulting performance and report it in the last section.

\section{Results}
\label{sec:results}
This section is divided into two subsections. First subsection \ref{subsec:results_full} shows the experimental comparison of our approach with two state-of-the art methods for near-isometric full shape matching. Section \ref{subsec:results_partial} demonstrates the effectiveness on partial shape matching. We evaluate all results by reporting the per-point-average geodesic distance between the ground truth map and the computed map. All results are multiplied by 100 for the sake of readability. 

\subsection{Full Shape Matching}\label{subsec:results_full} We present our results on a full shape matching benchmark dataset FAUST \cite{bogo2014}. This dataset contains $100$ shapes of $10$ different subjects in different poses where each point cloud contains $6890$ points. Following prior work, we use the last $20$ shapes as a test set and report the performance on this test set. We compare our results with \cite{ric_linear20, sharma20} in Table $2$ as they are applicable, in principle, to both partial and complete shape matching. Note that \cite{ric_linear20} presents results on the FAUST data that is subsampled to $1000$ points both during train and test. Our method obtains significantly better results than  \cite{ric_linear20}. Methods based on LBO eigen functions already form a good basis for shapes and thus, prior work based on LBO eigen basis obtains impressive performance. However, performance of this line of work degrades significantly under partiality, as shown in the next section and also in \cite{ric_linear20}.

\paragraph{Baselines} We compare with the following two state-of-the-art methods that are shown to outperform existing competitors and our ablated baselines: 

\textbf{\cite{sharma20}} This baseline, although weakly supervised, assumes to be given as input a mesh representation of a shape and LBO basis. We include it to demonstrate we can achieve competitive performance with the methods that excel in full shape matching where LBO basis are stable. 

\textbf{\cite{ric_linear20}} This is considered state of the art for learning embedding directly from data.  Since we are testing with point clouds of much higher resolution compared to the experiments in \cite{ric_linear20}, we retrain their models with $3k$ subsampled points for each point cloud and show their results with the best performing resolution of $3k$. The rest of the parameters such as embedding size are used as specified in their paper as they were found to be optimal. We use their open source code and retrain it on our subset of Surreal dataset. Note that this baseline is somewhat different from others since it requires and thus, learns both basis functions and probe functions (feature descriptors).

\textbf{Euc. Emb.} This baseline ablates the overall performance of our method and quantifies the gain brought in by the euclidean loss alone during training. It shows the performance if we learn an embedding by just projecting the shapes into a $50$ dimensional space with a nearest neighbour euclidean loss. 

\textbf{Euc. Emb. + comm.} This baseline combines the above baseline with the commutativity loss and quantifies what can be achieved without the linearly invariant assumption on self-symmetry map. We denote our results with \textbf{Euc. Emb. + comm. + Lin. Inv.} in the following Table. We also show the corresponding curves below that  are consistent with average geodesic error shown in the Table below. 

\begin{figure}[h]

   \begin{minipage}{0.49\textwidth}
\centering
\begin{tabular}{|l|c|}
\hline
Method $\backslash$ Dataset &  Faust  \\
\hline\hline
\cite{ric_linear20}-3k & 08\\
\cite{ric_linear20} + sym. & 09\\
Euc. Emb. & 12  \\
Euc. Emb. + comm. & 10 \\
Euc. Emb. + comm. + Lin. Inv. (Ours) & \textbf{05} \\
\cite{sharma20} & 05 \\
\hline
\end{tabular}
\label{table:res2}
    \end{minipage}%
    \begin{minipage}{0.49\textwidth}
       \centering
        \includegraphics[width=.85\linewidth]{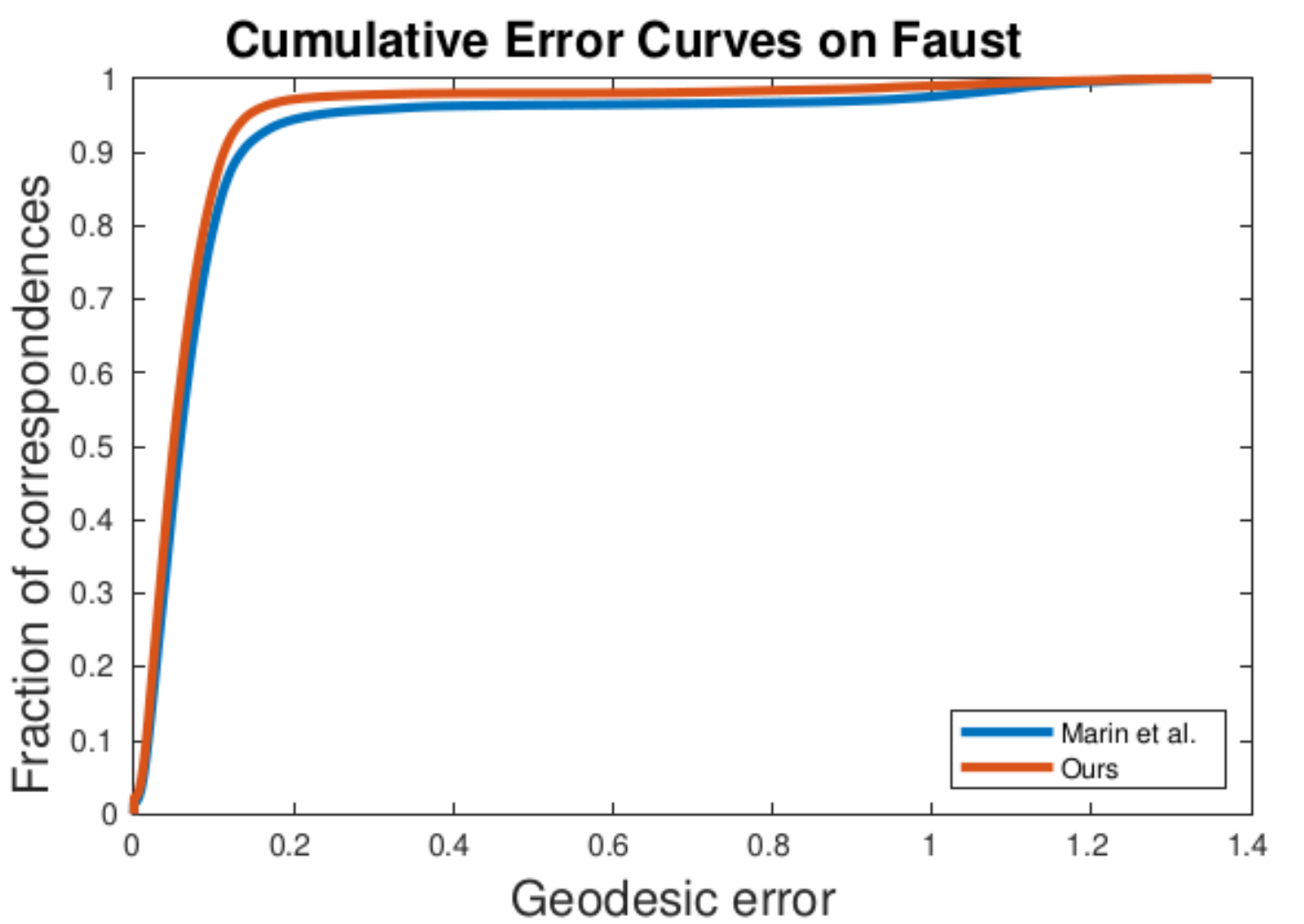} 
    \end{minipage}
    \label{fig:curves1}
\end{figure}

\paragraph{Discussion} Our ablation study shows that all three loss functions improve the overall performance. We note that the performance gains brought in by linearly invariant loss on self-symmetry embeddings are significant. We attribute our superior quantitative results over other learning based methods to a range of factors. First, in contrast to \cite{ric_linear20} that is based on two stage sequential architecture, our embedding is learned end to end in one phase. 
Second, none of the state-of-the-art methods takes symmetry structure into account even though symmetry ambiguities for shape matching is a well known problem and studied extensively in axiomatic methods. Third, performance of \cite{ric_linear20} is sensitive to the size of embedding $20$. In contrast, our method is quite robust and can train with twice their embedding size. We compete well with \cite{sharma20} even though this approach relies on LBO and exploiting mesh connectivity.

\subsection{Partial Shape Matching}
\label{subsec:results_partial}

\textbf{Datasets}
For a fair comparison with \citep{sharma20,litany17}, we follow the same experimental setup and test our method on the challenging SHREC’16 Partial Correspondence dataset \citep{cosmo2016matching}. The dataset is composed of
200 partial shapes, each containing about few hundreds to 9000 vertices, belonging to 8 different classes (humans and animals), undergoing nearly-isometric deformations in addition to having missing parts of various forms and sizes. Each class comes with a “null” shape in a standard pose which is used as the full template to which partial shapes are to be matched. The dataset is split into two sets, namely cuts (removal of a few large parts) and holes (removal of many small parts). We use the same test set following \citep{sharma20}. Overall, this test set contains $20$ shapes each for cuts and holes datasets chosen randomly from the two sets respectively. In addition to \cite{ric_linear20}, we compare with the following two baselines: \\

\textbf{\cite{sharma20}}. This baseline relies on learning LBO alignment and thus, is dependent on class that needs to be retrained for each of the 8 classes. We include their results even though our results are class agnostic and thus, significantly more robust and efficient. We obtain these results by running the code provided by the authors.

\textbf{\cite{litany17}}. This baseline is not learning based and relies on hand crafted features and an expensive optimization scheme on the Stiefel manifold for every pair of shapes at test time. Thus, in terms of computation and ground truth map requirement, it is most expensive.

\paragraph{Results and Discussion}
\label{subsec:quant_res}
We present our findings on partial shape matching in Table \ref{table:res1} where we obtain superior performance on both benchmark datasets for partial shape matching. In addition, our result outperforms the euclidean embedding competitive baseline by a significant margin and thus, validating the importance of working in the symmetric space while learning canonical embedding. We would like to stress that baseline such as \cite{sharma20} are class specific and need to be trained each time whereas our method is class agnostic and can obtain good results with a fraction of computational time. Similarly,  \cite{ric_linear20} trains a similar network as our two times. First, it learns an embedding with a network similar to ours, followed by a similar network training to compute the optimal linear transformation between the two embeddings. Moreover, the test phase also requires running the network twice. Therefore, our method is at least twice faster than this baseline in computational complexity.

\begin{table}

\caption{Avg. Geodesic Error on two partial SHREC benchmarks}
\begin{center}

\begin{tabular}{|l|c|c|}
\hline
Method $\backslash$ Dataset &  Holes& Cuts  \\
\hline\hline
\cite{litany17} & 16  &  13\\
 \cite{sharma20} & 14  &   16\\
\cite{ric_linear20}-3k & 12  & 15 \\
Euc. Emb. & 18  &  20\\
Ours & \textbf{10}  &   \textbf{13}\\
\hline
\end{tabular}
\end{center}
\label{table:res1}
\end{table}

\paragraph{Robustness to Embedding size} 
Our method is not sensitive to small variations in the embedding size. We demonstrate this by changing the embedding size and plotting the average error on the FAUST test set. Note that in contrast, \cite{ric_linear20} is extremely sensitive to changes in the size of feature descriptors (probe functions). We show the sensitivity in Figure $2$. 

\paragraph{Training Data Efficient}
We experiment with $3$ training set of different sizes $1000,3000, 5000$ sampled randomly and plot the results in Figure $3$. We observe a drop in the performance of our method but the drop is slightly better than our baseline \citep{ric_linear20}.
\begin{figure}[h]
\begin{minipage}{0.49\textwidth}
\centering
\caption{}
\centering
        \includegraphics[width=.9\linewidth]{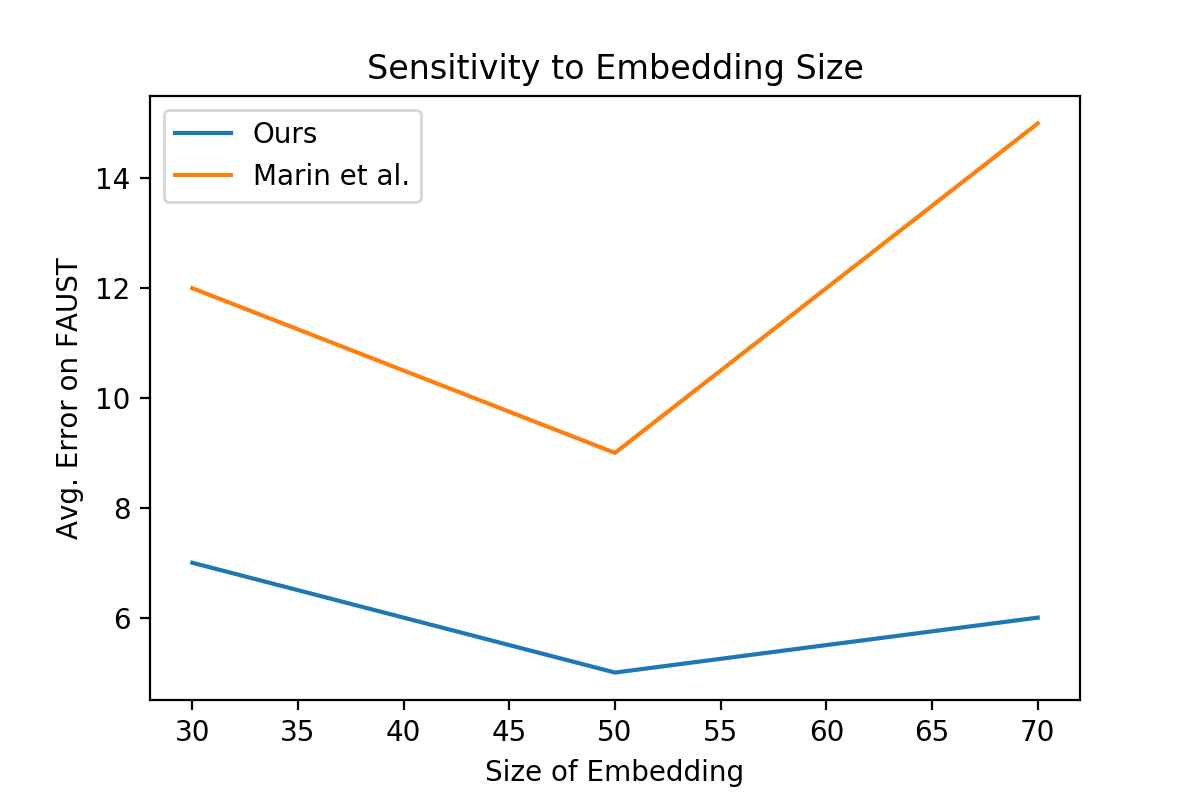} 
    \end{minipage}%
    \begin{minipage}{0.49\textwidth}
    \caption{\small{}}
       \centering
        \includegraphics[width=.9\linewidth]{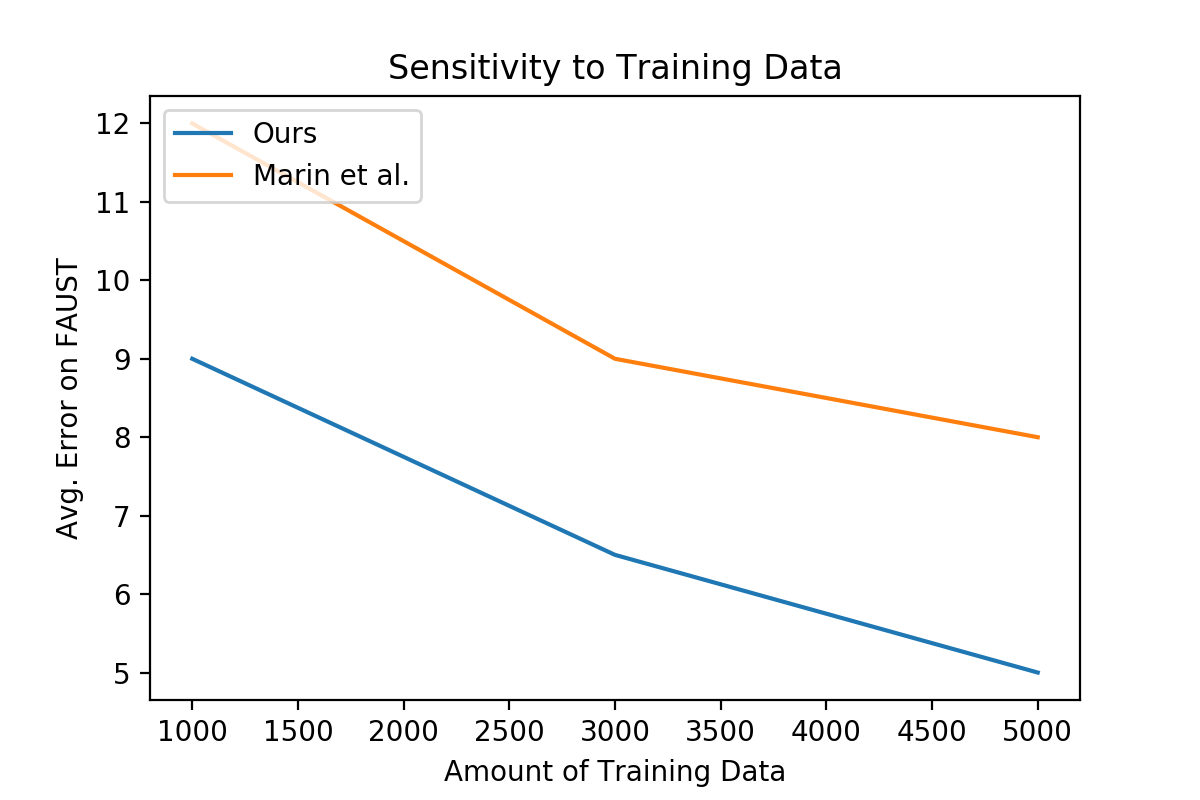} 
    \end{minipage}
    \label{fig:sensitive}
\end{figure}
\section{Conclusion and Future Work} In shape correspondence literature, partial shape matching and complete shape matching are generally tackled by two different sets of methods which obtain impressive results in one of the two respective domains. We presented a simple, general but effective method that reduces shape matching to a nearest neighbour search problem in a canonical embedding and apply it to both partial and complete shape matching. Our key idea is to learn an embedding of each shape that would make the given self-symmetry map linear in some higher-dimensional space. Our idea of injecting symmetry into the learning pipeline also serves as a regularizer and provides competitive performance on multiple shape matching benchmarks in comparison to all recent learning based methods.

There are several promising future directions to our work. First, our architecture is based on a very simple PointNet feature extractor and thus, there is a scope to integrate more advanced feature extractor that can exploit the neighbourhood structure better while learning the embedding. Second, following the advances in unsupervised deep functional maps, it would be interesting to explore learning of canonical embedding with minimal supervision. 

\section{Acknowledgement} Parts of this work were supported by the ERC Starting Grant StG-2017-758800 (EXPROTEA), and ANR AI chair AIGRETTE.

\bibliography{arxiv-main}
\bibliographystyle{iclr2022_conference}

\end{document}